\documentclass[preprint,12pt]{elsarticle}

\usepackage{amssymb}
\usepackage{amsmath}
\usepackage{booktabs}
\usepackage{algorithm}
\usepackage{algpseudocode}
\usepackage{multirow}

\newcommand{\thickrule}{\specialrule{0.1em}{0.25em}{0.25em}}

\algrenewcommand\algorithmicrequire{\textbf{Input:}}
\algrenewcommand\algorithmicensure{\textbf{Output:}}

\journal{Engineering Applications of Artificial Intelligence}

\begin{document}

\begin{frontmatter}

%% Title, authors and addresses

%% use the tnoteref command within \title for footnotes;
%% use the tnotetext command for theassociated footnote;
%% use the fnref command within \author or \affiliation for footnotes;
%% use the fntext command for theassociated footnote;
%% use the corref command within \author for corresponding author footnotes;
%% use the cortext command for theassociated footnote;
%% use the ead command for the email address,
%% and the form \ead[url] for the home page:
%% \title{Title\tnoteref{label1}}
%% \tnotetext[label1]{}
%% \author{Name\corref{cor1}\fnref{label2}}
%% \ead{email address}
%% \ead[url]{home page}
%% \fntext[label2]{}
%% \cortext[cor1]{}
%% \affiliation{organization={},
%%            addressline={}, 
%%            city={},
%%            postcode={}, 
%%            state={},
%%            country={}}
%% \fntext[label3]{}

\title{Dynamic Pricing in High-Speed Railways Using Multi-Agent Reinforcement Learning}

%% use optional labels to link authors explicitly to addresses:
%% \author[label1,label2]{}
%% \affiliation[label1]{organization={},
%%             addressline={},
%%             city={},
%%             postcode={},
%%             state={},
%%             country={}}
%%
%% \affiliation[label2]{organization={},
%%             addressline={},
%%             city={},
%%             postcode={},
%%             state={},
%%             country={}}

\author[affi1]{Enrique Adrian Villarrubia-Martin}
\ead{enrique.villarrubia@uclm.es}
\author[affi1]{Luis Rodriguez-Benitez\corref{cor1}}
\ead{luis.rodriguez@uclm.es}
\author[affi2]{David Muñoz-Valero}
\ead{david.munoz@uclm.es}
\author[affi3]{Giovanni Montana}
\ead{g.montana@warwick.ac.uk}
\author[affi1]{Luis Jimenez-Linares}
\ead{luis.jimenez@uclm.es}

%% Corresponding author
\cortext[cor1]{Corresponding author}

%% Author affiliation
\affiliation[affi1]{organization={Department of Technologies and Information Systems, Universidad de Castilla-La Mancha}, %Department and Organization
            addressline={Paseo de la Universidad 4}, 
            city={Ciudad Real},
            postcode={13071}, 
            %state={},
            country={Spain}}

\affiliation[affi2]{organization={Department of Technologies and Information Systems, Universidad de Castilla-La Mancha}, %Department and Organization
            addressline={Avenida Carlos III, s/n}, 
            city={Toledo},
            postcode={45071}, 
            %state={},
            country={Spain}}

\affiliation[affi3]{organization={Warwick Manufacturing Group, University of Warwick}, %Department and Organization
            addressline={Gibbet Hill Road}, 
            city={Coventry},
            postcode={CV4 7AL}, 
            %state={},
            country={UK}}

%% Abstract
\begin{abstract}
This paper addresses a critical challenge in the high-speed passenger railway industry: designing effective dynamic pricing strategies in the context of competing and cooperating operators. To address this, a multi-agent reinforcement learning (MARL) framework based on a non-zero-sum Markov game is proposed, incorporating random utility models to capture passenger decision making. Unlike prior studies in areas such as energy, airlines, and mobile networks, dynamic pricing for railway systems using deep reinforcement learning has received limited attention. A key contribution of this paper is a parametrisable and versatile reinforcement learning simulator designed to model a variety of railway network configurations and demand patterns while enabling realistic, microscopic modelling of user behaviour, called RailPricing-RL. This environment supports the proposed MARL framework, which models heterogeneous agents competing to maximise individual profits while fostering cooperative behaviour to synchronise connecting services. Experimental results validate the framework, demonstrating how user preferences affect MARL performance and how pricing policies influence passenger choices, utility, and overall system dynamics. This study provides a foundation for advancing dynamic pricing strategies in railway systems, aligning profitability with system-wide efficiency, and supporting future research on optimising pricing policies.
\end{abstract}

%%Graphical abstract
%\begin{graphicalabstract}
%\includegraphics{grabs}
%\end{graphicalabstract}

%%Research highlights
%\begin{highlights}
%\item A non-zero-sum Markov game models dynamic pricing strategies in railway systems.
%\item RailPricing-RL, a novel simulator, captures realistic railway networks and passenger behaviour.
%\item The multi-agent framework encourages cooperation between operators while maximising individual profitability.
%\item Results reveal the impact of dynamic pricing on passenger choice, utility, and system dynamics.
%\end{highlights}

%% Keywords
\begin{keyword}
%% keywords here, in the form: keyword \sep keyword

%% PACS codes here, in the form: \PACS code \sep code

%% MSC codes here, in the form: \MSC code \sep code
%% or \MSC[2008] code \sep code (2000 is the default)
Dynamic Pricing \sep Multi-Agent Reinforcement Learning \sep Social Dilemma \sep Discrete Choice Models \sep Railway Systems
\end{keyword}

\end{frontmatter}

%% Add \usepackage{lineno} before \begin{document} and uncomment 
%% following line to enable line numbers
%% \linenumbers

%% main text
%%

% --------------------
\section{Introduction}

Dynamic pricing is a pivotal strategy for revenue optimisation and demand management across industries. By adjusting prices in real time, based on supply and demand, companies can enhance profitability and better align services with consumer needs. However, there are important challenges involved in implementing effective dynamic pricing strategies, particularly in environments with complex interactions among multiple stakeholders. These challenges include modelling market dynamics, predicting consumer behaviour, and optimising pricing decisions, under constraints such as competition, cooperation, and limited resources. Machine learning (ML) has shown great promise in addressing these challenges, offering data-driven approaches to learning and adapting pricing strategies over time.

Reinforcement learning (RL), in particular, has emerged as a powerful tool for developing dynamic pricing strategies. RL enables agents to learn optimal policies through interactions with the environment, which means it is well suited to applications requiring real-time decision making. RL-based dynamic pricing has been successfully applied in sectors such as electricity markets \cite{Kim2015DynamicLearning,Qiu2020ALevels,Fraija2024DeepConstraints}, airlines \cite{Jo2024AirlineLearning}, and mobile networks \cite{Sun2023CompetitiveApproach}. These studies primarily focus on single-agent and multi-agent RL settings, where agents aim to maximise profits through individual or collaborative actions. Despite this progress, dynamic pricing for high-speed rail systems remains under-researched, even though the industry poses unique challenges and opportunities compared to other domains.

\begin{figure}[t]
    \centering
    \includegraphics[width=0.8\textwidth]{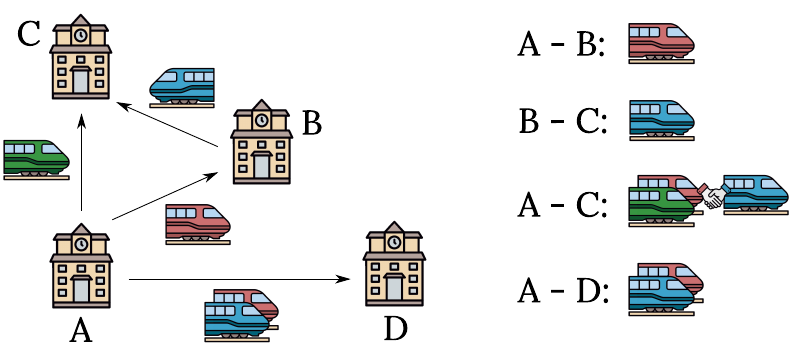}
       \caption{Illustration of a railway network model, where stations are represented as nodes and connections as edges. Edges are colour-coded to indicate the companies operating services between stations. The A-B and B-C markets are operated by separate companies, but through cooperation, they can offer connecting services in the A-C market, which otherwise would have a direct service from only one company. Additionally, the blue and red companies compete directly in the A-D market. Best viewed in colour.}
    \label{fig:introduction}
\end{figure}

High-speed railways are characterised by an intricate interplay of competition and cooperation between operators. Figure~\ref{fig:introduction} illustrates a simplified railway network, where stations are nodes and connections are edges representing services provided by different companies. Operators may cooperate to offer connecting journeys (e.g., A-C via A-B and B-C) or compete directly on shared routes (e.g., A-D). These interactions require sophisticated pricing strategies, as decisions in one market can ripple through others. Moreover, as networks grow larger and more interconnected, operators must balance profitability with seamless service integration, further complicating the optimisation of pricing strategies.

In order to apply RL effectively in this context, it is crucial to model passenger decision making accurately, as this directly affects the environment in which agents learn. Discrete Choice Modelling (DCM) is a widely used tool to model and predict travel choices. These models, grounded in the principles of classic Random Utility Models (RUMs), provide a framework for understanding how travellers make decisions from among a finite set of alternatives. DCM assigns utility values to travel options, assuming individuals choose the option with the highest utility. While DCM has been effectively applied to urban railway systems \cite{Koh2017SimulationEnvironment}, these applications typically focus on short-distance, high-frequency travel. In contrast, intercity railways require models capable of handling a range of passenger behaviours, multi-operator networks, and longer travel distances. Despite their utility, existing simulators such as ROBIN \cite{delCastillo-Herrera2024ROBIN:SimulatioN} are limited in their ability to model dynamic pricing and connecting services, and they lack compatibility with RL frameworks like OpenAI Gymnasium \cite{Towers2024Gymnasium:Environments}. These limitations lead to gaps in the ability to develop and evaluate advanced RL-based dynamic pricing strategies for high-speed railways.

To address these gaps, this paper introduces a multi-agent reinforcement learning (MARL) framework for dynamic pricing in high-speed railway networks. Central to this framework is a novel RL simulator called RailPricing-RL that extends ROBIN by enabling dynamic pricing, modelling multi-operator journeys, and supporting MARL algorithms. The simulator creates a mixed cooperative-competitive environment where heterogeneous agents dynamically adjust ticket prices in response to demand fluctuations. This framework allows for the study of pricing strategies to balance competition and cooperation, capturing the unique dynamics of the high-speed railway domain.

Through extensive experiments, the proposed framework is evaluated by testing advanced MARL algorithms, such as Multi-Actor Attention Critic (MAAC) \cite{Iqbal2019Actor-attention-criticLearning} and Multi-Agent Deep Deterministic Policy Gradient (MADDPG) \cite{Lowe2017Multi-agentEnvironments}, in the context of dynamic pricing for high-speed rail networks. These experiments explore how agents adapt to mixed cooperative-competitive dynamics and study the effects of user preferences on agent performance, equity, and system-wide outcomes under varying demand scenarios. The results highlight the algorithms' ability to balance trade-offs between cooperation and competition, optimise pricing strategies, and align individual profitability with broader system efficiency. Furthermore, the findings of the study reveal the challenges posed by heterogeneous agent interactions and the significant role of user preferences in shaping overall system behaviour, offering key insights for the design of robust MARL-based solutions for real-world applications.

The remainder of this paper is organised as follows: Section \ref{sec:related_work} reviews related work, including dynamic pricing, Deep Reinforcement Learning (DRL) applications in railways, social dilemmas in MARL, and related RL environments. Section \ref{sec:preliminaries} formalises the mixed multi-agent task as a stochastic game. Section \ref{sec:methods} details the proposed methodology, and Section \ref{sec:experiments} presents experimental results. Finally, Section \ref{sec:conclusions} concludes the paper and discusses future lines of research.

% --------------------
\section{Related work} \label{sec:related_work}

To contextualise the contributions, this section reviews prior research in four key areas. First, dynamic pricing strategies (Section \ref{sec:dynamic_pricing}) are explored, as they provide the foundation for optimising prices in high-speed railway systems. Second, the use of DRL in railway systems (Section \ref{sec:deep_rl}) is examined, highlighting its role in addressing operational challenges. Third, social dilemmas in multi-agent systems (Section \ref{sec:dilemmas}) are discussed, as they capture the interplay between cooperative and competitive behaviours, central to the proposed framework. Finally, a review of related RL environments (Section \ref{sec:related_environments}) positions the contributions within the broader context of MARL research and simulation platforms.

\subsection{Dynamic pricing} \label{sec:dynamic_pricing}

Dynamic pricing involves adjusting prices in real-time based on supply and demand to maximise profits and align market conditions. This concept, also known as pricing intelligence, has been widely explored in various fields through DRL methods.

In electricity markets, DRL has been used to optimise prices in energy consumption and transportation. For example, \cite{Kim2015DynamicLearning} developed algorithms enabling suppliers and customers to learn dynamic pricing and consumption scheduling without prior knowledge, reducing system costs. Similarly, for Electric Vehicle (EV) charging, \cite{Qiu2020ALevels} combined the Deep Deterministic Policy Gradient (DDPG) \cite{Lillicrap2016ContinuousLearning} algorithm with Prioritised Experience Replay (PER) \cite{Schaul2016PrioritizedReplay} to enhance pricing strategies in deregulated systems. For Hydrogen Fuel Cell Vehicle (HFCEV) refuelling, \cite{Fraija2024DeepConstraints} used DRL to coordinate refuelling schedules and determine prices, improving demand satisfaction and traffic flow in microgrids.

Dynamic pricing has also been applied in telecommunications. \cite{Sun2023CompetitiveApproach} presented a model for interactions between Mobile Virtual Network Operators (MVNOs) and users, framing the problem as a Stackelberg game and solving it using a Multi-Agent Deep Q-Network (MADQN). In smart grids, \cite{Ma2024OptimalAlgorithm} proposed a distributed multi-agent optimisation approach for resolving supply-demand imbalances, strengthening privacy, autonomy, and efficiency in dynamic pricing. In air transport, dynamic pricing shares similarities with the rail sector. \cite{Jo2024AirlineLearning} used DRL to optimise pricing for patient customers, demonstrating that alternating between high and low prices during a sales period can improve revenues. These findings are particularly relevant for industries such as railways, where customer preferences and temporal dynamics significantly influence pricing strategies.

In contrast to these domains, there are further special challenges in dynamic pricing in high-speed railways. Operators must account for network interdependencies, requiring the synchronisation of multi-operator journeys, while navigating overlapping markets where pricing decisions in one market can ripple through to others due to connecting services. Furthermore, the competitive-cooperative dynamics in railways requires frameworks that can handle heterogeneous agents and multi-operator networks. Existing methods, such as those applied in electricity, telecommunications, and air transport, do not adequately address these challenges, as they often assume homogeneous agents, independent markets, or simpler interactions.

\subsection{DRL in railway systems} \label{sec:deep_rl}

The use of DRL in railway systems has shown significant potential for optimising operations and improving efficiency. Key applications include rail traffic optimisation \cite{Ning2022DeepOptimization}, where DRL generates recommended trajectories for trains in real time, ensuring punctuality and energy efficiency. Train Timetable Rescheduling (TTR) \cite{Yue2024ReinforcementRepresentation} and the Vehicle Rescheduling Problem (VRSP) \cite{Mohanty2020Flatland-rl:Trains} use DRL to restore operations quickly after disruptions by adjusting train schedules or reassigning vehicles to routes.

Other applications include predictive maintenance scheduling \cite{Mohammadi2022APlanning,Arcieri2024POMDPMaintenance}, energy management, autonomous train control \cite{Feng2021ANetwork,Cui2020Knowledge-basedRailway}, and supply chain optimisation for goods transportation. Simulation environments such as Flatland-RL \cite{Mohanty2020Flatland-rl:Trains} play a crucial role in enabling the safe, efficient development and testing of DRL algorithms. For example, Flatland-RL simplifies complex railway tasks like VRSP, while reducing computational costs, as demonstrated by the NeurIPS 2020 Benchmark \cite{Li2021ScalableChallenge}. For train interval control, \cite{Lin2024TrackingLearning} applied secure DRL to balance safety and traffic density by using vehicle-to-vehicle communication and Constrained Markov Decision Processes (CMDPs). Their approach improved safety by 30\% while increasing system efficiency.

However, most existing applications of DRL in railways focus on operational challenges, such as scheduling and maintenance, rather than market-driven strategies like dynamic pricing. Addressing dynamic pricing requires additional capabilities, such as modelling passenger behaviour, capturing interdependencies in multi-operator journeys, and balancing competition and cooperation between operators.

\subsection{Social dilemmas in MARL} \label{sec:dilemmas}

Social dilemmas in MARL arise when agents' decisions influence not only their own rewards but also those of others. These dilemmas are critical in scenarios where cooperation and competition coexist, requiring strategies that balance individual incentives with system-wide outcomes. In the context of MARL, addressing social dilemmas involves designing mechanisms and environments that encourage cooperation when beneficial while allowing agents to compete effectively.

\cite{Leibo2017Multi-agentDilemmas} introduces sequential social dilemmas, extending the concept of cooperation and competition from isolated actions to temporally extended policies rather than isolated actions, as seen in traditional matrix games like the prisoner's dilemma. The study demonstrates how environmental factors, such as resource abundance, shape agent behaviour and lead to conflicts over shared resources. Building on this, \cite{Hughes2018InequityDilemmas} explores inequity aversion as a mechanism for promoting fairness and cooperation, while \cite{Jaques2019SocialLearning} introduces causal influence rewards to enhance coordination in MARL settings. These studies highlight how carefully designed reward structures can incentivise cooperative behaviour even in competitive environments.

More recent research has focused on decentralised decision making and localised cooperation in MARL. For instance, \cite{Yi2022LearningLearning} proposes a hierarchical framework where agents dynamically share rewards with neighbouring agents, fostering cooperation at a local level while pursuing broader objectives. Similarly, \cite{Xing2024ARules} considers mechanisms to sustain stable cooperation by adjusting agent strategies based on their experiences. Another work, \cite{Sun2025GDT:Space}, introduces adaptive grouping through group mesh topology, enabling agents with high similarity to share parameters while maintaining individual value functions and overcoming monotonicity constraints. These approaches underscore the importance of dynamic policies that adapt to both cooperative and competitive contexts.

The proposed MARL framework reflects many of these principles in its design. Social dilemmas emerge naturally in the environment, particularly in the trade-offs between cooperative behaviours, such as synchronising connecting services, and competitive pricing strategies in overlapping markets. Unlike traditional team-based MARL settings, the agents in this framework operate without fixed teams, and cooperation provides localised benefits specific to the participating agents, rather than system-wide rewards. This dynamic aligns with real-world scenarios, where agents optimise individual profits while navigating cooperative opportunities and competitive constraints.

The framework builds on the idea of mixed cooperation and competition as explored in \cite{Leibo2017Multi-agentDilemmas}. It extends it, however, to the domain-specific context of high-speed railways. By enabling agents to dynamically interact based on localised incentives, the framework offers a unique perspective on how social dilemmas manifest in real-world multi-agent environments. Through the integration of random utility models and a flexible simulator, the framework provides a foundation for studying the interplay between individual incentives and collective outcomes in complex MARL settings.

\subsection{Related RL environments} \label{sec:related_environments}

In the literature, existing RL environments often model mixed settings where agents pursue individual goals while cooperating in specific contexts. For example, in the \textit{Clean Up} task from the Melting Pot \cite{Leibo2021ScalablePot} evaluation suite, agents must balance harvesting berries for individual rewards with cleaning a river for the common good. Similarly, in the \textit{Predator and Prey} task from the Multi Particle Environment \cite{Mordatch2018EmergencePopulations,Lowe2017Multi-agentEnvironments}, predators must cooperate to catch faster prey, as rewards depend on capturing the prey but not on the number of participating predators.

Other environments, such as \textit{Google Research Football} \cite{Kurach2020GoogleEnvironment}, Melting Pot's \textit{Capture the Flag}, and variations of \textit{Predator and Prey}, focus on team-based settings where agents cooperate within fixed teams while competing against others. In contrast, the proposed environment features no fixed teams: agents can simultaneously compete and collaborate with the same opponents. Cooperation occurs in specific contexts, such as multi-operator journeys, and benefits only the participating agents rather than providing a shared system-wide advantage. This design captures the dynamics of real-world railway networks, where agents optimise individual objectives while navigating both cooperative and competitive market interactions.

% --------------------
\section{Preliminaries} \label{sec:preliminaries}

This study considers a mixed multi-agent task that can be modelled as a stochastic game $G$, which satisfies the Markov property. More specifically, $G$ is a Markov Game \cite{Littman1994MarkovLearning}, a multi-agent extension of Markov Decision Processes (MDPs). The game is defined by the tuple $G = \langle S, U, P, r, Z, O, n, \gamma \rangle$. Here, $S$ represents the set of possible states, with $s \in S$ denoting the current state of the environment. The $n$ agents, denoted by $a \in A \equiv \{1, \dots, n\}$, select actions $u^a \in U$ at each time step. These actions collectively form a joint action $\mathbf{u} \in \mathbf{U} \equiv U^n$, which induces a transition to a new state according to the state transition function $P(s'|s, \mathbf{u}) : S \times \mathbf{U} \times S \to [0, 1]$. 

Each agent has an individual reward function, defined as $r^a(s, \mathbf{u}) : S \times \mathbf{U} \to \mathbb{R}$, which depends on the global state and the joint action. This is in contrast to Decentralised Partially Observable Markov Decision Processes (Dec-POMDPs) \cite{Oliehoek2016APOMDPs}, typically used in fully cooperative settings where all agents share the same reward function. Stochastic games, however, allow for both competitive and cooperative interactions between agents.

In the partially observable setting considered here, each agent receives an observation $z^a \in Z$, which provides partial information about the global state. The observation is determined by the observation function $O(s, a) : S \times A \to Z$. Each agent aims to learn a stochastic policy $\pi^a(u^a|z^a) : Z \times U \to [0, 1]$, which maps observations to a probability distribution over actions. The objective of each agent is to maximise its expected discounted return, defined as:

\begin{equation*}
    J(\pi^a) = \mathbb{E}_{\mathbf{u} \sim \boldsymbol{\pi}, s \sim P} \left[ \sum_{t=0}^\infty \gamma^t r_t^a(s_t, \mathbf{u}_t) \right],
\end{equation*}

where $\gamma \in [0, 1]$ is the discount factor that determines the relative importance of immediate versus long-term rewards.

% --------------------
\section{Methods} \label{sec:methods}

This section presents the methodology used to simulate and analyse the dynamic pricing problem in high-speed railway networks. The proposed framework incorporates a journey-based simulator to model passenger decision making, operator-defined services, and demand patterns across multiple markets (Section \ref{sec:journey}). Furthermore, Section \ref{sec:environment} describes the key components of the environment, including the observation space, action space, and reward function used in the MARL framework.

\subsection{Journey-based high-speed railway simulation} \label{sec:journey}

\begin{figure}[t]
    \centering
    \includegraphics[width=0.8\textwidth]{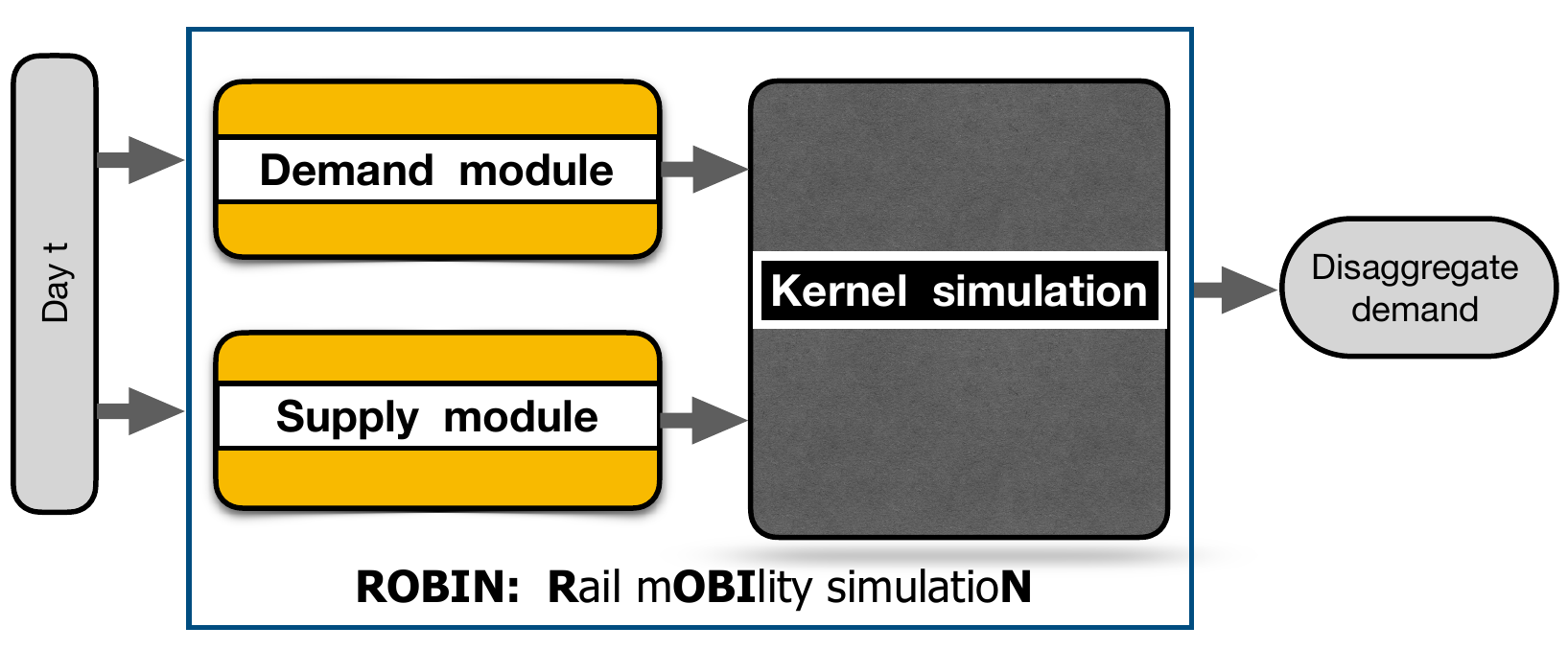}
    \caption{Architecture of the ROBIN simulator, comprising the supply module, demand module, and simulation kernel. The kernel integrates daily operator-defined services and probabilistically-generated demand across origin-destination markets to simulate high-speed railway dynamics.}
    \label{fig:robin}
\end{figure}

The simulation environment is based on the ROBIN simulator, which generates microscopic interactions between supply and demand for passengers in high-speed railway networks. It comprises three main components: a supply module, a demand module, and a simulation kernel integrating these elements (Figure~\ref{fig:robin}).

The supply module defines the set of railway services provided by operators, including schedules, routes, and ticket prices. The demand module generates daily passenger demand by simulating user preferences, travel patterns, and market-level variations. The simulation kernel integrates these components, modelling interactions between passenger decisions and operator strategies, such as pricing adjustments and use of the services.

Unlike the original service-oriented ROBIN simulator, the proposed framework adopts a journey-based model. This enables multi-operator journeys and dynamic interactions to be modelled, better reflecting real-world travel dynamics. By structuring the simulation environment in this way, the framework captures the complexities of interconnected railway networks and provides a flexible platform for evaluating dynamic pricing strategies.

\paragraph{Demand modelling}

Service demand for a given day $t$ is represented as:

\begin{equation} \label{eq:demand}
    {\cal D}_t = \left ( \{d_{\omega }\}_{\omega \in W},  \{p^\omega_k\}_{k\in {\cal K}}  \right )_p,
\end{equation}

where $W$ denotes the set of markets (origin-destination pairs), and ${\cal K}$ represents the set of passenger types. The term $\{d_\omega\}_{\omega \in W}$ represents the potential demand for each market $\omega$, parameterised as random variables accounting for day-to-day variations. The term $\{p^\omega_k\}_{k \in {\cal K}}$ specifies the probability distribution over passenger types $k$ in market $\omega$, indicating the likelihood of passengers belonging to each type.

The demand module takes daily samples from these distributions to generate randomised demand inputs, ensuring that variations in market conditions and passenger behaviour are realistically modelled.

\paragraph{Journey-based model}

A journey $J$ represents a passenger's complete travel itinerary across origin-destination pairs. It is defined as $J = (S, w, {\cal \tau}, t')$, where $S = \{s_1, s_2, ..., s_n\}$ is an ordered sequence of services where each $s_i \in S$ for $i \in \{1, 2, ..., n\}$ is an individual service operated by a railway company, $w = (w_o, w_d)$ specifies the market with $w_o$ and $w_d$ denoting the stations of origin and destination, ${\cal \tau} = \{t_{s_1}, t_{s_2}, ..., t_{s_n}\}$ represents the departure times for the services in $S$, and $t'$ is the travel date.

A journey is valid if it satisfies specific conditions. Each consecutive pair of services $(s_i, s_{i+1})$ must have a transfer time ($t_{s_{i+1}} - t_{s_i}$) greater than or equal to the minimum transfer time ($\delta_{\min}$). The journey must also align with service schedules and connect the desired origin and destination.

Table~\ref{tab:journeys} shows examples of valid and invalid journeys in the A-C market. Valid journeys meet all the required criteria, including minimum transfer times and sequential connectivity. Invalid journeys fail to satisfy one or more of these constraints. Transfer times are recorded as ``-" when no transfer is required.

\begin{table}[!t]
    \caption{Examples of valid and invalid journeys in the A-C market with $\delta_{min} = 5$ minutes at day $t'$. Valid journeys meet all conditions for scheduling and transfer times, while invalid journeys violate at least one of these constraints. Transfer times of ``-" indicate no transfer is required.}
    \scriptsize
    \begin{tabular*}{\textwidth}{@{\extracolsep\fill}cccccc}
        \toprule%
        \multicolumn{6}{@{}c@{}}{\textbf{Valid Journeys}} \\\midrule%
        Journey & Service & Market & Departure Time & Arrival Time & Transfer Time \\
        \midrule
        $J_1$ & $s_1$ & (A, C) & 08:00 & 09:00 & - \\
        \midrule
        $J_2$ & $s_2$ & (A, B) & 08:00 & 08:45 & - \\
        $J_2$ & $s_3$ & (B, C) & 09:00 & 09:30 & 00:15 \\
        \thickrule
    \end{tabular*}
    \begin{tabular*}{\textwidth}{@{\extracolsep\fill}cccccc}
        \multicolumn{6}{@{}c@{}}{\textbf{Invalid Journeys}} \\\midrule%
        Journey & Service & Market & Departure Time & Arrival Time & Transfer Time \\
        \midrule
        $J_3$ & $s_4$ & (A, B) & 08:00 & 08:50 & - \\
        $J_3$ & $s_5$ & (B, D) & 09:00 & 09:25 & 00:10 \\
        \midrule
        $J_4$ & $s_6$ & (A, B) & 08:00 & 08:30 & - \\
        $J_4$ & $s_7$ & (B, D) & 08:50 & 09:15 & 00:20 \\
        $J_4$ & $s_8$ & (D, C) & 09:15 & 09:45 & 00:00 \\
        \bottomrule
    \end{tabular*}
    %\footnotetext{}
    \label{tab:journeys}
\end{table}

\paragraph{Metrics for journey evaluation}

The structure and efficiency of a journey are quantified using two key metrics. The first is the number of transfers, denoted by $N_{\text{transfers}}$, which measures the total number of service changes within the journey. For a journey comprising $n$ services, this is calculated as:

\begin{equation*}
    N_{\text{transfers}} = n - 1.
\end{equation*}

This metric provides insight into the complexity of the journey. Fewer transfers indicate a more straightforward and potentially more convenient journey for passengers.

The second metric is the total transfer time, denoted as $T_{\text{transfer}}$, which measures the cumulative time passengers spend transferring between services. It is defined as:

\begin{equation*}
    T_{\text{transfer}} = \sum_{i=1}^{n-1} \left( t_{s_{i+1}} - t_{s_i} \right),
\end{equation*}

where $t_{s_i}$ and $t_{s_{i+1}}$ represent the arrival and departure times of consecutive services, respectively. This metric captures the aggregate waiting time at transfer stations, which directly impacts passenger experience and satisfaction. A lower $T_{\text{transfer}}$ indicates a more efficient journey with minimal delays between connecting services.

\paragraph{Passenger utility modelling}

Passenger behaviour is represented using a utility function which captures the factors influencing their journey choices:

\begin{align}
    U_{jcnt'} =  \xi_{TSP,k} + \delta_{ck} - f_k(AT_{jk}) - r_k(DT_{jk}) - g_k(P_{c\omega t'}) \nonumber \\
    - h_k(TT_{j\omega}) - \tau_k(TR_{j\omega}) - \ell_k(NT_{j\omega}) + {\rm rand} (\varepsilon_{k}),
    \label{eq:utility}
\end{align}

where $U_{jcnt'}$ represents the utility of a specific journey $j$ for passenger $n$ of type $k$, given a seat $c$ on day $t'$. Each term in the equation reflects a specific aspect of the passenger’s decision making. For instance, $\xi_{TSP,k}$ captures the passenger's perception of the train service provider, while $\delta_{ck}$ refers to the perceived utility of the seat and $g_k(P_{c\omega t'})$ accounts for  sensitivity to ticket price. Other components, such as $f_k(AT_{jk})$ and $r_k(DT_{jk})$, model preferences for arrival and departure times, respectively.

The utility function also includes factors such as the total travel time ($h_k(TT_{j\omega})$), transfer time ($\tau_k(TR_{j\omega})$), and the number of transfers ($\ell_k(NT_{j\omega})$). A random error term, ${\rm rand} (\varepsilon_{k})$, accounts for unobserved variability in passenger preferences. In addition, for journeys involving multiple seats and railway companies, the average utility of the components $\xi_{TSP,k} + \delta_{ck}$ is calculated, and if at least one service within the journey has no available tickets, the utility of a journey is set to $-\infty$.

Passengers select the journey $j^*$ that maximises their utility, provided that utility is positive:
\begin{equation} \label{eq:best_journey}
    j^* = \arg\max_{j \in J} U_{jcnt'} > 0.
\end{equation}
If all available journeys yield a utility of zero or less, the passenger chooses not to travel. This decision-making process reflects how passengers assess trade-offs between price, travel time, convenience, and other factors when selecting a journey.

\paragraph{Simulation transition dynamics}

The simulation models the interactions between passenger decisions and operator strategies, capturing the dynamics of demand generation, journey selection, and ticket purchasing. At the start of each simulated day, the supply and demand modules are initialised to define the available services and generate passenger demand $\mathcal{D}_t$, based on market conditions and user patterns. Valid journeys $\mathcal{J}$ are then filtered to ensure they meet scheduling constraints and the minimum transfer time $\delta_{\text{min}}$.

Passenger decision making is modelled using a utility-based approach. For each journey, the simulation evaluates service availability and seat quality, computing utility values at seat level that account for factors such as price, travel time, and passenger preferences. Passengers select the journey that maximises their utility, provided it is positive; otherwise, they opt not to travel. The system iterates this process, dynamically adjusting the state of the environment as passenger choices and operator strategies evolve. The detailed transition dynamics are formalised in \ref{sec:transition_dynamics}.

\paragraph{Strategic interactions}

The proposed framework models a non-zero-sum game where passengers can opt out if no journey meets their utility threshold. This feature captures the realistic scenario in which passenger demand is elastic, influenced by operators' pricing strategies and service quality. By allowing for the emergence of new travellers or the loss of potential passengers based on these strategies, the framework adds complexity to operator interactions. This dynamic interplay reflects the real-world trade-offs between competitive pricing and service attractiveness, shaping the overall market dynamics.

\subsection{Design of the dynamic pricing environment} \label{sec:environment}

This section outlines the implementation of the Markov game introduced earlier in the context of the MARL framework for dynamic pricing in high-speed railway networks. The environment is designed to simulate interactions between agents (operators) and the system, providing a platform to study their strategies under realistic conditions. The key components of the environment include the observation space (Section  \ref{sec:observation_space}), which defines the information available to each agent, the action space (Section \ref{sec:action_space}), which specifies the decisions agents can make, and the reward function (Section \ref{sec:reward}), which quantifies the outcomes of their strategies.

\subsubsection{Observation space} \label{sec:observation_space}

The observation space defines the information accessible to agents in the MARL framework, capturing the state of the railway system in order to support dynamic pricing decisions. Each observation $O_s$ corresponds to a specific service $s \in {\cal S}$, where ${\cal S}$ denotes the set of all services in the supply module. Observations include both static and dynamic attributes.

Static attributes represent fixed properties of a service, such as the train service provider, the corridor (a set of stations within a region), the line (a sequence of stations in the corridor), the time slot, and the rolling stock. These are encoded as integer indices for computational efficiency. Dynamic attributes, on the other hand, capture real-time service information, including pricing and ticket sales. Pricing data specifies the departure and arrival stations and the seat prices for each seat type in an origin-destination pair. Ticket sales data reflects the number of seats sold for each seat type, providing a snapshot of service demand.

In the MARL context, agents observe only a subset of the full observation space, tailored to their role as operators. Public information, such as train service providers, prices, and time slots, is available to all agents. However, private data, such as ticket sales, is restricted to services operated by the corresponding agent. For services not operated by an agent, the tickets sold attribute is excluded. Formally, the observation space for an agent $a$ is given by:

\begin{equation*}
    Z_a = \left\langle
    \begin{cases}
        O_s \setminus \{\text{tickets\_sold}\}, & \text{if } TSP(s) \neq a \\
        O_s, & \text{otherwise}
    \end{cases},
    \forall s \in {\cal S} \right\rangle
\end{equation*}

where $TSP(s)$ denotes the train service provider operating service $s$.

This design ensures that agents receive the information necessary for decision making while maintaining privacy for sensitive data. The initial observation space is fixed and configurable based on the demand hyperparameters, allowing flexibility for different simulation scenarios.

\subsubsection{Action space} \label{sec:action_space}

The action space defines the set of decisions available to agents for modifying ticket prices in their respective services. Each action represents a percentage change in the price of available seats for a specific service in the supply module. Rather than setting prices directly, agents adjust them relative to their current values, which aligns with real-world pricing practices.

Formally, if the current price for a seat type $c$ in a market $w$ at time step $t$ is $p_{wt}^c$, and the action specifies a percentage change $\alpha$, the updated price at time step $t+1$ is:

\begin{equation*}
    p_{wt+1}^c = p_{wt}^c \cdot \left(1 + \alpha \cdot \frac{\beta}{100}\right),
\end{equation*}

where $\alpha \in [-1, 1]$ is a normalised value representing a percentage increase ($\alpha > 0$) or decrease ($\alpha < 0$), and $\beta$ is a scaling factor set to 25 by default. Prices are clipped to ensure they remain within a valid range $[0, \infty)$, preventing negative pricing.

The environment also supports a discrete action space for scenarios where algorithms require fixed action sets. In this case, each action corresponds to one of eleven discrete values, representing five levels of price increase, five levels of price decrease, and an option to leave the price unchanged. The relative modification of prices, rather than absolute adjustments, offers several benefits. It reflects common dynamic pricing practices, where prices are adjusted incrementally based on market conditions \cite{Kopalle2023DynamicDirections}, and reduces the risk of destabilising the system with abrupt changes, which can negatively impact training convergence and performance.

\subsubsection{Reward} \label{sec:reward}

At each time step $t$, measured in days, an agent $a$ receives a reward $r^a_t$ which quantifies the profit generated from the tickets sold for its services. The reward is calculated as the difference in total profit between the current time step, denoted by $\rho_t^s$, and the previous time step, $\rho_{t-1}^s$:

\begin{equation*}
    r^a_t = \sum_{s \in S^a} \rho_t^s - \rho_{t-1}^s,
    \label{eq:reward}
\end{equation*}

where $S^a$ is the set of services operated by agent $a$. This formulation assumes no associated costs for providing services, simplifying the reward function and focusing solely on revenue changes.

The reward structure encourages agents to maximise short-term profits by aligning their strategies to immediate revenue gains. However, this dense reward formulation may incentivise aggressive tactics such as predatory pricing \cite{Rosenthal1981GamesParadox}, where agents drastically lower ticket prices to attract customers from competitors, potentially leading to long-term inefficiencies. 

Over the course of an episode, the total return for an agent $a$ is the discounted sum of its rewards:
\begin{equation*}
    R_t^a = \sum_{l=0}^T \gamma^l r_{t+l}^a,
\end{equation*}
where $T$ is the terminal time step of the episode, and $\gamma \in [0,1]$ is the discount factor, controlling the trade-off between immediate and long-term rewards. This reward mechanism incentivises agents to continuously refine their pricing strategies to optimise profitability throughout the simulation.

% --------------------
\section{Experimental settings and results} \label{sec:experiments}

This section evaluates the proposed MARL framework for dynamic pricing in high-speed railway networks. The scenarios used for the evaluation are described in Section~\ref{sec:scenarios}, the algorithms considered are outlined in Section~\ref{sec:algorithms}, and the details of implementation are given in Section~\ref{sec:implementation_details}. The simulation framework provides a flexible platform to explore a range of research questions, such as:

\begin{itemize}
    \item How do different MARL algorithms perform in terms of \textit{profitability}, \textit{equity}, and \textit{adaptability} across varying user demand patterns?
    \item To what extent do agent pricing policies influence \textit{passenger behaviour}, including travel decisions, utility, and the total number of passengers?
    \item How do \textit{cooperative} and \textit{competitive dynamics} shape agent strategies and system-wide outcomes? Can mechanisms such as attention or shared rewards enhance performance in mixed-agent environments?
\end{itemize}

The answers to these research questions are provided in Section~\ref{sec:performance_analysis}, which presents the performance comparison and analysis of the algorithms in the proposed framework, and in Section~\ref{sec:further_studies}, which explores additional insights from further studies. These experiments highlight the adaptability of the framework to diverse scenarios, uncovering trade-offs between profitability, equity, and cooperation. The insights thus gained contribute to broader MARL challenges, such as learning in non-stationary environments, and managing heterogeneous agent objectives.

\subsection{Scenarios} \label{sec:scenarios}

The experiments use a fixed rail network topology as described in the introduction, where stations are nodes and connections are edges operated by different companies. The design incorporates both competitive and cooperative dynamics, with agents competing in shared markets and collaborating to provide connecting services. Two key markets, A-C and A-D, capture direct and connecting journeys. The agents are heterogeneous, as their action spaces differ and all services use a single seat type to focus the analysis on dynamic pricing strategies. Specifically, the characteristics of the scenarios are described as follows.

\textbf{Business.} This scenario models a single user group with inelastic demand. Passengers are relatively insensitive to price changes, simulating a stable market dominated by business travellers. Each episode spans 5 days, with an average of 110 passengers across all markets.

\textbf{Business \& Student.} This scenario introduces two user groups with distinct price sensitivities. Business travellers are less price-sensitive and more likely to book tickets in advance, while students are highly price-sensitive and tend to buy tickets closer to their travel dates. Episodes last 7 days, with an average of 220 passengers across all markets, comprising 60\% business and 40\% student travellers.

These scenarios provide a structured framework for analysing agent behaviour, user preference, and system-wide outcomes. By comparing results across the scenarios, the experiments demonstrate the MARL framework's ability to model dynamic pricing challenges in both simple and complex market conditions.

\subsection{Algorithms for evaluation} \label{sec:algorithms}

To evaluate the performance of the proposed MARL framework and its applicability to dynamic pricing, a diverse set of algorithms from both single-agent and multi-agent paradigms is tested. These algorithms are chosen to reflect a range of strategies and learning dynamics, allowing for a comprehensive analysis of agent behaviour, system outcomes, and pricing strategies in both cooperative and competitive scenarios. 

The inclusion of single-agent RL serves as a benchmark to understand the performance of algorithms in a simplified monopolistic setting, where a single agent controls all services without interference from competitors. This allows the impact of competition and cooperation introduced in the multi-agent setting to be isolated. The selected single-agent RL algorithms are:

\begin{itemize}
    \item \textbf{TD3 \cite{Fujimoto2018AddressingMethods}:} A deterministic actor-critic algorithm designed for continuous action spaces. TD3 is included for its robustness in stabilising learning through delayed policy updates and target networks, making it a strong choice for settings with high-dimensional action spaces.
    \item \textbf{SAC \cite{Haarnoja2018SoftActor}:} A stochastic policy gradient algorithm that balances reward maximisation with exploration by optimising an entropy-augmented objective. SAC is included for its ability to handle exploration-exploitation trade-offs effectively, which is critical in dynamic environments.
\end{itemize}

In the MARL setting, agents operate in a shared environment, interacting through both cooperative and competitive dynamics. The following MARL algorithms are included to explore different approaches to learning in mixed-agent environments:

\begin{itemize}
    \item \textbf{IQL-SAC \cite{Tan1993Multi-AgentAgents,Haarnoja2018SoftActor}:} A decentralised approach where each agent learns independently, treating others as part of the environment. Despite the inherent non-stationarity caused by changing policies of other agents, IQL-SAC provides a baseline for independent learning in MARL.
    \item \textbf{VDN-SAC \cite{Sunehag2017Value-DecompositionLearning,Haarnoja2018SoftActor}:} A cooperative algorithm that combines individual value functions into a shared global value. VDN-SAC is included to evaluate how shared rewards influence agent cooperation, despite their limited applicability in competitive real-world scenarios.
    \item \textbf{MAAC \cite{Iqbal2019Actor-attention-criticLearning}:} A multi-agent extension of SAC incorporating attention mechanisms. MAAC is selected for its ability to focus on relevant agent interactions, which is particularly useful in mixed cooperative-competitive settings.
    \item \textbf{MADDPG \cite{Lowe2017Multi-agentEnvironments}:} An extension of DDPG with centralised training and decentralised execution. MADDPG is included to evaluate its performance in environments with explicit cooperation and competition.
\end{itemize}

Finally, random policies are included to provide a point of comparison against non-learning, purely stochastic decision-making strategies. 

These algorithms are tested across the Business and Business \& Student scenarios described in Section~\ref{sec:scenarios}, enabling a systematic evaluation of their strengths and limitations in modelling dynamic pricing strategies and agent interactions in high-speed railway networks.

\subsection{Implementation details} \label{sec:implementation_details}

For a standardised comparison, the default hyperparameters of each algorithm were used, without additional tuning. A complete list of hyperparameters is provided in \ref{sec:hyperparameters}. All experiments were run on an Intel i7-13700KF CPU and an NVIDIA GeForce RTX 4080 16GB GPU.

To encourage initial exploration and diversify the state-action space, each algorithm was first trained with a random policy for 1,000 episodes. A replay buffer of one million steps was used to store experience for all algorithms, and each model was trained for 200,000 episodes. For algorithms such as TD3, SAC, IQL-SAC, and VDN-SAC, delayed policy updates were performed with a frequency of two for every Q-network update.

To ensure robust results, experiments were conducted with 16 parallel environments, each initialised with a unique random seed. Additionally, training and testing were conducted with separate sets of random seeds to avoid overfitting and ensure generalisability. Each experiment was repeated with three independent seeds, providing reliable statistical estimates for performance metrics.

Reward normalisation was applied during training to stabilise learning and improve convergence. The normalised reward $\hat{r}_t$ at time step $t$ was computed as:

\begin{equation*}
    \hat{r}_t = \frac{r_t - \mu_t}{\sqrt{\sigma_t^2 + \epsilon}},
\end{equation*}

where $r_t$ is the raw reward at time $t$, $\mu_t$ is the running mean of rewards up to time $t$, $\sigma_t^2$ is the running variance of rewards up to time $t$, and $\epsilon$ is a small constant for numerical stability.

\subsection{Performance comparisons and analysis} \label{sec:performance_analysis}

This section analyses the performance of various algorithms under single-agent and multi-agent settings across both scenarios. Table~\ref{tab:total_profits} summarises the total profits obtained during evaluation.

In the \textbf{single-agent setting}, agents optimise their policies independently, without considering the presence of competitors. Both TD3 and SAC outperform the random policy baseline in the Business and Business \& Student scenarios, with TD3 achieving the highest performance in both cases. These results highlight the effectiveness of these algorithms in simpler, monopolistic environments where agents focus solely on maximising their own rewards.

The \textbf{multi-agent setting} introduces significant complexity as agents must account for the actions of others, leading to both collaborative and competitive dynamics. In the Business scenario, all multi-agent algorithms perform better than the random policy. However, in the more complex Business \& Student scenario, characterised by diverse user patterns, most algorithms, with the exception of VDN-SAC and MAAC, struggle to outperform the random policy. VDN-SAC achieves the best performance due to its use of shared rewards, which encourage cooperation between agents. However, shared rewards are often impractical in real-world settings due to legal and regulatory constraints on pricing coordination. On the other hand, MAAC benefits from its attention mechanism, allowing it to adapt to dynamic inter-agent interactions, particularly in scenarios with heterogeneous passenger behaviours.

\begin{table}[!t]
    \caption{Total profits obtained at evaluation for the Business and Business \& Student scenarios by the algorithms. The mean and the standard deviation are calculated by three independent runs.}
    \scriptsize
    \begin{tabular*}{\textwidth}{@{\extracolsep\fill}lcccccc}
        \toprule%
        & \multicolumn{3}{@{}c@{}}{Single-Agent} \\\cmidrule{2-4}%
        Scenario  & Random & TD3 & SAC \\
        \midrule
        Business  & $5061\pm130$ & $\mathbf{11081\pm100}$ & $9319\pm66$ \\
        Business \& Student  & $8626\pm39$& $\mathbf{13403\pm246}$ & $10520\pm195$  \\
        \thickrule
    \end{tabular*}
    \begin{tabular*}{\textwidth}{@{\extracolsep\fill}lcccccc}
        & \multicolumn{5}{@{}c@{}}{Multi-Agent} \\\cmidrule{2-6}%
        Scenario  & Random & IQL-SAC & VDN-SAC* & MAAC & MADDPG \\
        \midrule
        Business  & $5061\pm130$ & $\mathbf{6487\pm414}$ & $\mathbf{8127\pm393}$ & $5636\pm159$ & $6433\pm3446$\\
        Business \& Student  & $8626\pm39$ & $7884\pm94$  & $\mathbf{10268\pm29}$ & $\mathbf{9098\pm2918}$ & $6517\pm2310$\\
        \bottomrule
    \end{tabular*}
    The highest scores are indicated in bold. In the Multi-Agent setting, the highest scores for algorithms that do not use shared rewards are also highlighted in bold.
    \label{tab:total_profits}
\end{table}

\begin{figure}[!t]
    \centering
    \includegraphics[width=\textwidth]{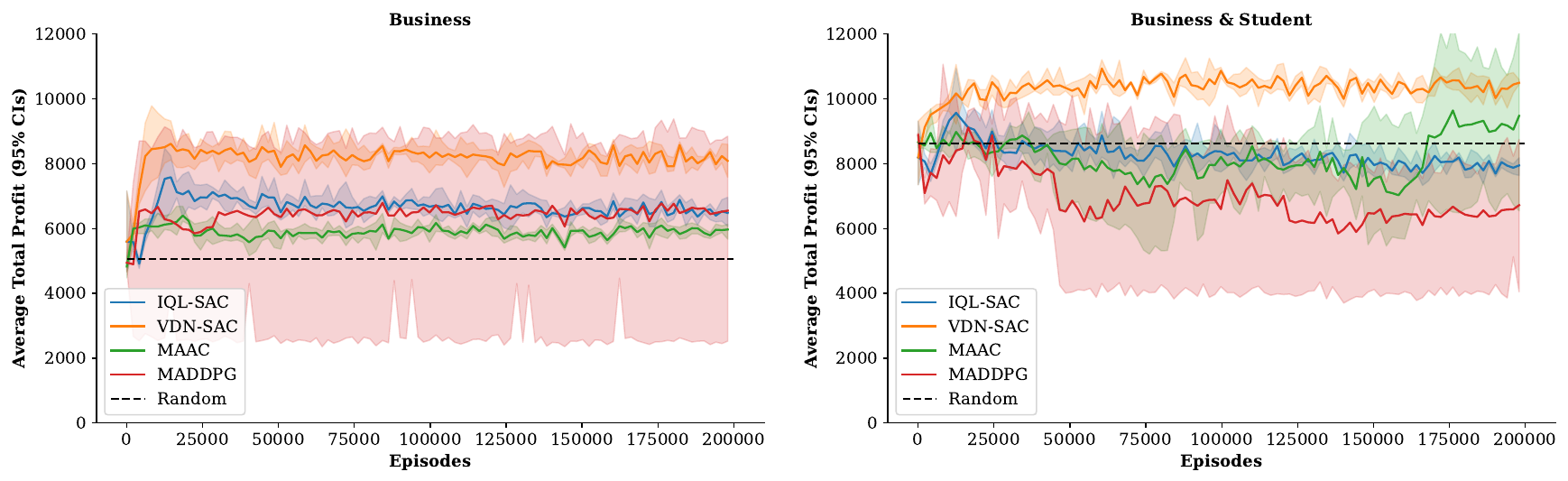}
    \caption{Average total profits obtained at training for the Business and Business \& Student scenarios by the algorithms. Best viewed in colour.}
    \label{fig:training_curves}
\end{figure}

The training dynamics, shown in Figure~\ref{fig:training_curves}, provide additional insights into the performance of these algorithms. In the Business scenario, most algorithms stabilise after an initial improvement phase, suggesting that the simpler dynamics of this scenario are easier to learn. In contrast, the training dynamics for the Business \& Student scenario reveal greater variability. Algorithms such as IQL-SAC, MADDPG, and even MAAC initially perform poorly and exhibit performance degradation over time. This decline reflects the increased difficulty of managing the diverse user patterns and inter-agent interactions in this more complex scenario. However, MAAC shows late-stage improvement due to its ability to focus on key agent interactions through its attention mechanism, as discussed in Section~\ref{sec:attention}.

These results underscore the novelty and complexity of the dynamic pricing problem. Traditional MARL algorithms often struggle to maintain stable performance and even perform better than the random policy in environments with diverse user patterns and complex inter-agent dynamics. This highlights the need for further advancements in MARL techniques to handle such challenges effectively.

\subsection{Further studies} \label{sec:further_studies}

This section considers various aspects of agent behaviour and its impact on the environment, offering a deeper analysis of key dynamics within the proposed framework. The analysis in Section~\ref{sec:equality} assesses the equality of profit distribution between agents, identifying algorithms that achieve more balanced outcomes. This highlights the trade-offs between individual profitability and overall equity in multi-agent systems. Section~\ref{sec:passengers} examines the relationship between agent pricing policies and passenger behaviour, focusing on their effects on utility and the percentage of passengers who choose to travel. These findings reveal the tension between maximising agent profits and ensuring passenger satisfaction.

Section~\ref{sec:attention} explores the role of the attention mechanism in the MAAC algorithm, particularly in complex scenarios with heterogeneous interactions between agents and passengers. The analysis demonstrates how attention mechanisms can improve agent performance by prioritising relevant interactions. Finally, Section~\ref{sec:cooperation_competition} studies the interplay between cooperation and competition between agents, showing how a strategic balance between these dynamics can improve system-wide outcomes.

\subsubsection{How equally distributed are the profits among the agents?} \label{sec:equality}

The equality of profit distribution between agents serves as a key metric for assessing the fairness of policies learned by different algorithms. This is particularly relevant in scenarios where agents share access to a common-pool resource. Equality is quantified using the metric proposed in \cite{Perolat2017AAppropriation}, defined as:

\begin{equation*}
    E = 1 - \frac{\sum_{i=1}^N \sum_{j=1}^N \lvert R^i - R^j \rvert}{2N \sum_{i=1}^N R^i},
\end{equation*}

where $E$ represents the equality metric, $R^i$ and $R^j$ are the profits of agents $i$ and $j$ at the end of the episode, and $N$ is the total number of agents. Higher values of $E$ indicate a more equitable distribution of profits. Profits are used instead of returns to ensure a fair comparison with VDN-SAC, which uses shared rewards across agents.

\begin{table}[!t]
    \caption{Equality at evaluation for the Business and Business \& Student scenarios by the algorithms. The mean and the standard deviation are calculated by three independent runs.}
    \scriptsize
    \begin{tabular*}{\textwidth}{@{\extracolsep\fill}lcccccc}
        \toprule%
        & \multicolumn{3}{@{}c@{}}{Single-Agent} \\\cmidrule{2-4}%
        Scenario  & Random & TD3 & SAC \\
        \midrule
        Business  & $0.705\pm0.00$ & $\mathbf{0.725\pm0.00}$ & $0.706\pm0.01$ \\
        Business \& Student  & $0.777\pm0.00$& $0.728\pm0.00$ & $\mathbf{0.881\pm0.02}$  \\
        \thickrule
    \end{tabular*}
    \begin{tabular*}{\textwidth}{@{\extracolsep\fill}lcccccc}
        & \multicolumn{5}{@{}c@{}}{Multi-Agent} \\\cmidrule{2-6}%
        Scenario  & Random & IQL-SAC & VDN-SAC* & MAAC & MADDPG \\
        \midrule
        Business  & $0.705\pm0.00$ & $0.685\pm0.01$ & $0.733\pm0.00$ & $0.625\pm0.02$ & $\mathbf{0.773\pm0.09}$\\
        Business \& Student  & $0.777\pm0.00$ & $0.756\pm0.01$  & $\mathbf{0.897\pm0.01}$ & $0.702\pm0.14$ & $0.719\pm0.09$\\
        \bottomrule
    \end{tabular*}
    The highest scores are indicated in bold.
    \label{tab:equality}
\end{table}

Table~\ref{tab:equality} presents the equality metrics for the Business and Business \& Student scenarios. Of all the algorithms, VDN-SAC achieves the highest equality in both scenarios, due to its shared reward structure that inherently promotes cooperation between agents. However, this approach is less applicable in real-world settings, where such coordination is typically unfeasible due to legal constraints. In contrast, other algorithms, including MAAC and MADDPG, exhibit lower equality, particularly in the more complex Business \& Student scenario, where heterogeneous user patterns make equitable profit distribution more challenging.

\begin{figure}[!t]
    \centering
    \includegraphics[width=\textwidth]{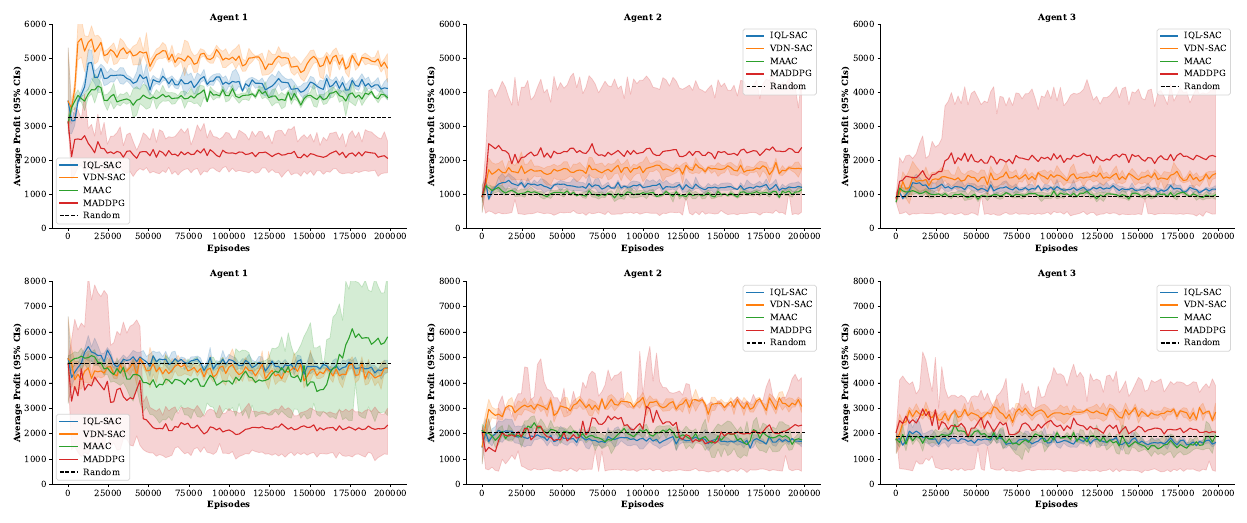}
    \caption{Average total profits earned by agents at training for the Business and Business \& Student scenarios by the algorithms. Best viewed in colour.}
    \label{fig:agents_profit}
\end{figure}

Figure~\ref{fig:agents_profit} provides additional insights into individual profits during training. In the Business scenario, MADDPG achieves nearly equal profits across agents, with Agents 2 and 3 performing particularly well. However, this result is influenced by the significantly lower performance of Agent 1, which reduces competitive pressure and allows the other agents to excel. In the Business \& Student scenario, MAAC shows improved performance, primarily driven by Agent 1. While Agents 2 and 3 achieve profits comparable to those of the random policy, the disparity in performance across agents highlights the difficulty of achieving equity in multi-agent settings with a range of passenger behaviours.

These results underscore the complexity of balancing profitability and fairness in MARL environments. While shared rewards can improve equality, they may not be practical in competitive real-world scenarios, necessitating further research into alternative strategies that promote equity without undermining competitiveness.

\subsubsection{How does the policy of the agents affect passenger utility and travel decisions?} \label{sec:passengers}

This section examines how the pricing policies learned by the agents influence passenger utility, travel decisions, and system inclusivity. Tables~\ref{tab:utility} and~\ref{tab:travelling} illustrate the relationship between agent profits and passenger outcomes over the scenarios evaluated.

The results show a clear trade-off between agent profitability and passenger utility. Algorithms that achieve higher profits often result in lower passenger utility due to increased ticket prices. For instance, in the Business \& Student scenario, TD3 maximises profits but reduces passenger utility to $3.15 \pm 0.05$ (Table~\ref{tab:utility}), as high prices limit affordable options for passengers. This highlights the inherent tension between economic goals and passenger satisfaction.

\begin{table}[!t]
    \caption{Passenger utility at evaluation for the Business and Business \& Student scenarios by the baselines. The mean and the standard deviation are calculated by three independent runs.}
    \scriptsize
    \begin{tabular*}{\textwidth}{@{\extracolsep\fill}lcccccc}
        \toprule%
        & \multicolumn{3}{@{}c@{}}{Single-Agent} \\\cmidrule{2-4}%
        Scenario  & Random & TD3 & SAC \\
        \midrule
        Business  & $7.94\pm0.01$ & $\mathbf{5.24\pm0.02}$ & $6.08\pm0.02$ \\
        Business \& Student  & $6.24\pm0.12$& $\mathbf{3.15\pm0.05}$ & $5.30\pm0.34$  \\
        \thickrule
    \end{tabular*}
    \begin{tabular*}{\textwidth}{@{\extracolsep\fill}lcccccc}
        & \multicolumn{5}{@{}c@{}}{Multi-Agent} \\\cmidrule{2-6}%
        Scenario  & Random & IQL-SAC & VDN-SAC* & MAAC & MADDPG \\
        \midrule
        Business  & $7.94\pm0.0$ & $7.31\pm0.14$ & $\mathbf{6.51\pm0.12}$ & $7.68\pm0.06$ & $7.00\pm1.90$\\
        Business \& Student  & $6.24\pm0.12$ & $8.41\pm0.06$  & $\mathbf{5.27\pm0.13}$ & $6.33\pm3.37$ & $7.89\pm3.54$\\
        \bottomrule
    \end{tabular*}
    The lowest scores are indicated in bold.
    \label{tab:utility}
\end{table}

Moreover, TD3 achieves its profitability by targeting business travellers, who are more willing to pay higher ticket prices. However, this strategy excludes price-sensitive groups, such as students. Table~\ref{tab:travelling} shows that only $59.99 \pm 0.69$\% of passengers choose to travel in the Business \& Student scenario under TD3, with no student passengers included. Such outcomes demonstrate the risks of non-inclusive pricing policies, which undermine sustainable transportation objectives by prioritising economic gains over equity.

In contrast, multi-agent algorithms such as MAAC and VDN-SAC achieve more balanced results in scenarios with heterogeneous user preferences. For example, MAAC achieves moderate utility and travel rates in the Business \& Student scenario, reflecting its ability to adapt pricing strategies dynamically. However, the variability in outcomes across algorithms underscores the complexity of balancing profitability, passenger utility, and inclusivity in multi-agent environments.

These findings emphasise the importance of designing policies that account for diverse passenger preferences, aligning economic, social, and sustainability objectives in real-world transportation systems.

\begin{table}[!t]
    \caption{Percentage of passengers travelling at evaluation for the Business and Business \& Student scenarios by the algorithms. The mean and the standard deviation are calculated by three independent runs.}
    \scriptsize
    \begin{tabular*}{\textwidth}{@{\extracolsep\fill}lcccccc}
        \toprule%
        & \multicolumn{3}{@{}c@{}}{Single-Agent} \\\cmidrule{2-4}%
        Scenario  & Random & TD3 & SAC \\
        \midrule
        Business  & $100\pm0.00$ & $\mathbf{99.62\pm0.01}$ & $99.99\pm0.02$ \\
        Business \& Student  & $91.10\pm0.86$& $\mathbf{59.99\pm0.69}$ & $89.47\pm6.73$  \\
        \thickrule
    \end{tabular*}
    \begin{tabular*}{\textwidth}{@{\extracolsep\fill}lcccccc}
        & \multicolumn{5}{@{}c@{}}{Multi-Agent} \\\cmidrule{2-6}%
        Scenario  & Random & IQL-SAC & VDN-SAC* & MAAC & MADDPG \\
        \midrule
        Business  & $\mathbf{100\pm0.00}$ & $\mathbf{100\pm0.00}$ & $\mathbf{100\pm0.00}$ & $\mathbf{100\pm0.00}$ & $\mathbf{100\pm0.00}$\\
        Business \& Student  & $91.10\pm0.86$ & $99.93\pm0.01$  & $89.28\pm1.32$ & $\mathbf{79.60\pm17.67}$ & $86.22\pm15.39$\\
        \bottomrule
    \end{tabular*}
    The lowest scores are indicated in bold.
    \label{tab:travelling}
\end{table}

\subsubsection{Does the attention mechanism used by the MAAC algorithm influence its performance?} \label{sec:attention}

\begin{figure}[!t]
    \centering
    \includegraphics[width=\textwidth]{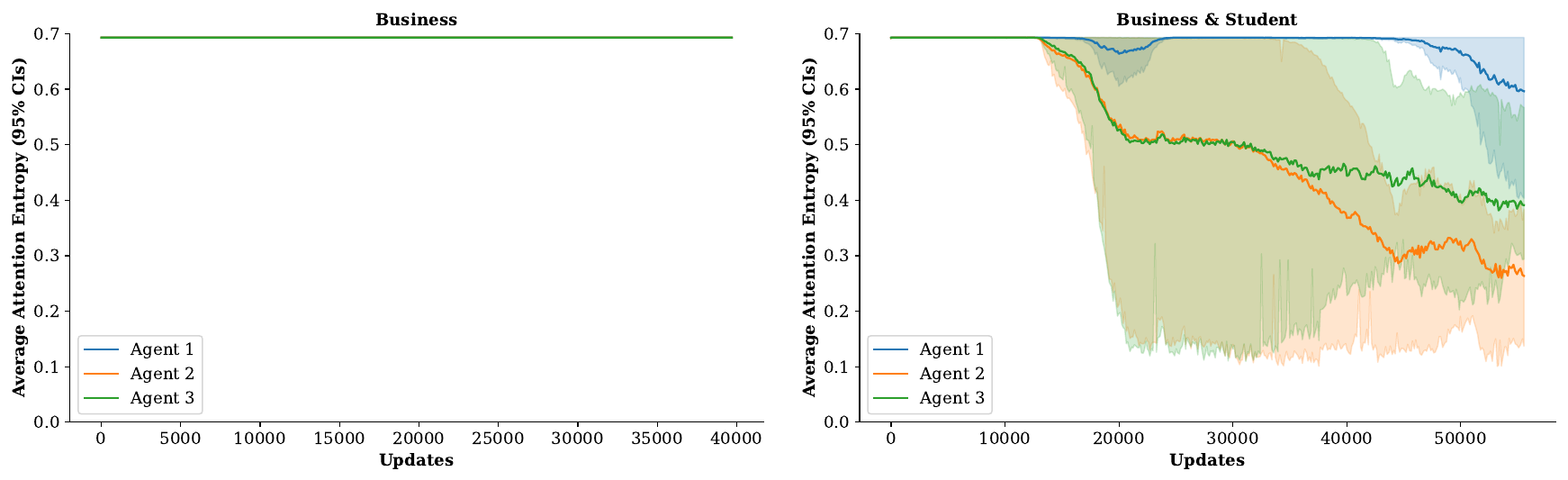}
    \caption{Average attention entropy per agent during training for the Business and Business \& Student scenarios using the MAAC algorithm. Best viewed in colour.}
    \label{fig:attention_entropy}
\end{figure}

To evaluate the impact of the attention mechanism in MAAC, Figure~\ref{fig:attention_entropy} illustrates the average attention entropy during training for each agent using four attention heads, computed according to Eq.~(\ref{eq:entropy}). Attention entropy quantifies the distribution of attention weights across agents, with lower entropy indicating more focused attention. An entropy of zero signifies complete focus on a single agent, while higher entropy reflects a uniform distribution of attention.

\begin{equation} \label{eq:entropy}
    H_i = \frac{1}{K} \sum_{k=1}^K \left( -\frac{1}{T} \sum_{t=1}^T \sum_{j=1}^N \alpha_{i,j,t}^k \cdot \log(\alpha_{i,j,t}^k + \epsilon) \right)
\end{equation}

Here, $H_i$ is the average entropy for agent $i$, $K$ is the total number of attention heads per agent, $T$ is the number of time steps, $N$ is the number of agents attended to, $\alpha_{i,j,t}^k$ is the attention weight for agent $i$ and head $k$ focusing on agent $j$ at time $t$, and $\epsilon$ is a small constant for numerical stability.

In the Business scenario, the attention mechanism fails to identify meaningful interactions, resulting in uniformly distributed attention weights across all agents. This lack of focus corresponds to suboptimal performance, as agents are unable to prioritise relevant interactions effectively.

In contrast, in the more complex Business \& Student scenario, MAAC achieves performance improvements over the random policy due to its ability to adaptively focus attention. As training progresses, agents dynamically assign higher attention weights to specific peers. Notably, the performance improvement is primarily driven by Agent 1, as evidenced by the decrease in its attention entropy, highlighting the importance of focusing on relevant interactions to manage heterogeneous user patterns effectively.

\subsubsection{Is cooperating and competing between agents better than just competing?} \label{sec:cooperation_competition}

To evaluate the impact of cooperation and competition on total profits, the policy distributions of the MADDPG algorithm are analysed under two different seeds. As shown in Table~\ref{tab:distribution}, the behaviour and performance of agents vary significantly depending on whether cooperation emerges alongside competition.

In the first scenario, agents operate in a purely competitive dynamic with no cooperation. Here, agents primarily focus on reducing ticket prices across all markets to attract more passengers, as indicated by the high probability of selecting actions that produce maximum price reductions. While this strategy increases individual passenger volumes, it significantly reduces overall ticket prices, resulting in an average total profit of $2,503$. These findings demonstrate that pure competition undermines the profitability of the system by driving prices to unsustainable levels.

In the second scenario, a mix of cooperation and competition emerges. Cooperation is evident in markets such as A-B (Agent 2) and B-C (Agent 3), where Agent 2 lowers prices while Agent 3 raises them. From the passenger's perspective, the total ticket cost over the connecting journey remains balanced. In contrast, markets like A-C and C-D exhibit competitive behaviour, where agents occasionally increase prices, as reflected in the probabilities of actions that lead to maximum price increases. This strategic balance between cooperation and competition results in a significantly higher average total profit of $7,856$, demonstrating the potential benefits of coordinated yet competitive agent interactions.

These results highlight the importance of fostering strategic cooperation alongside competition in multi-agent environments. By enabling agents to balance their pricing strategies across markets, the system can achieve greater profitability while maintaining competitive dynamics.

\begin{table}[!t]
    \caption{Policy distribution at evaluation for the Business and Business \& Student scenarios by the MADDPG algorithm. The purely competing case corresponds to seed = 0, while the cooperating and competing case corresponds to seed = 43.}
    \scriptsize
    \begin{tabular*}{\textwidth}{@{\extracolsep\fill}lcccccc}
        \toprule
        \multicolumn{7}{@{}c@{}}{\textbf{Competing (Total Average Profit: 2503)}} \\
        & \multicolumn{2}{@{}c@{}}{Agent 1} & \multicolumn{2}{@{}c@{}}{Agent 2} & \multicolumn{2}{@{}c@{}}{Agent 3} \\\cmidrule{2-3}\cmidrule{4-5}\cmidrule{6-7}%
        Action/Market & A - C & - & A - B & C - D & B - C & C - D \\
        \midrule
        Maximum price reduction & \textbf{0.2} & - & \textbf{0.2} & \textbf{0.2} & \textbf{0.2} & \textbf{0.2} \\
        Moderate price reduction & 0 & - & 0 & 0 & 0 & 0 \\
        No price adjustment & \textbf{0.8} & - & \textbf{0.8} & \textbf{0.8} & \textbf{0.8} & \textbf{0.8} \\
        Moderate price increase & 0 & - & 0 & 0 & 0 & 0 \\
        Maximum price increase & 0 & - & 0 & 0 & 0 & 0 \\
        \thickrule
    \end{tabular*}
    \begin{tabular*}{\textwidth}{@{\extracolsep\fill}lcccccc}
        \multicolumn{7}{@{}c@{}}{\textbf{Cooperating and Competing (Total Average Profit: 7856)}} \\
        & \multicolumn{2}{@{}c@{}}{Agent 1} & \multicolumn{2}{@{}c@{}}{Agent 2} & \multicolumn{2}{@{}c@{}}{Agent 3} \\\cmidrule{2-3}\cmidrule{4-5}\cmidrule{6-7}%
        Action/Market & A - C & - & A - B & C - D & B - C & C - D \\
        \midrule
        Maximum price reduction & 0.0 & - & \textbf{0.2} & 0.0 & 0.0 & 0.0 \\
        Moderate price reduction & 0 & - & 0 & 0 & 0 & 0 \\
        No price adjustment & \textbf{0.8} & - & \textbf{0.8} & \textbf{0.8} & \textbf{0.8} & \textbf{0.8} \\
        Moderate price increase & 0 & - & 0 & 0 & 0 & 0 \\
        Maximum price increase & \textbf{0.2} & - & 0 & \textbf{0.2} & \textbf{0.2} & \textbf{0.2} \\
        \bottomrule
    \end{tabular*}
    The probabilities that are not equal to zero are indicated in bold.
    \label{tab:distribution}
\end{table}

% --------------------
\section{Conclusions and future work} \label{sec:conclusions}

This study proposes a novel MARL framework for addressing dynamic pricing in high-speed railway networks, with several key contributions. At the core of the work is a novel RL simulator called RailPricing-RL that extends the ROBIN framework by enabling dynamic pricing, modelling multi-operator journeys, and supporting MARL algorithms. This simulator provides a flexible platform for evaluating agent behaviour and pricing strategies in complex railway networks, incorporating both competitive and cooperative dynamics across direct and connecting services. Additionally, the study introduces scenarios with distinct user demand patterns to analyse the adaptability of MARL algorithms to varying levels of price sensitivity and market complexity.

The experimental results demonstrated the challenges and opportunities associated with applying MARL to dynamic pricing. Shared-reward approaches, such as VDN-SAC, returned the best equality outcomes, but remain impractical in real-world settings due to regulatory constraints on pricing coordination. Attention-based mechanisms, such as those in MAAC, showed promise in managing mixed-agent interactions by dynamically balancing competition and cooperation. However, traditional MARL algorithms often struggled to handle the nuanced dynamics of the system, particularly in scenarios with diverse user preferences. Aggressive pricing strategies maximised profits but reduced passenger utility and inclusivity, highlighting the need for sustainable pricing policies that align economic and social objectives.

Future research should focus on expanding the capabilities of the RL simulator and the MARL framework. Incorporating more complex network topologies with additional markets and services would enable a deeper exploration of the interplay between competition and cooperation in dynamic pricing. Furthermore, there is a need to develop MARL algorithms tailored to this domain, with strategies that explicitly promote fairness and long-term sustainability while maintaining robust performance. Recent advances in value decomposition, such as sequence value decomposition transformers \cite{Zhao2025SequenceLearning}, offer promising directions by considering unequal agent interactions through action sequences, which could better capture the asymmetric nature of railway operator relationships where some agents have greater market influence than others.

Finally, extending the reward formulation to include cost functions and operator constraints would further enhance the simulation’s real-world applicability, offering richer insights into the challenges of dynamic pricing in high-speed railway systems. Integrating operational considerations such as infrastructure reliability could provide additional depth to pricing strategies, where techniques from sensor optimization and fault diagnosis \cite{Wang2025SensorLearning} could inform dynamic pricing adjustments based on system health and maintenance requirements. Such integration would offer richer insights into the challenges of balancing economic objectives with operational constraints in high-speed railway systems.

% --------------------
\section*{CRediT authorship contribution statement}
\textbf{Enrique Adrian Villarrubia-Martin:} Writing – original draft, Visualisation, Validation, Software, Methodology, Data curation, Conceptualisation. \textbf{Luis Rodriguez-Benitez:} Writing – original draft, Resources, Methodology,  Formal analysis. \textbf{David Muñoz-Valero:} Writing – review \& editing, Data curation, Software.  \textbf{Giovanni Montana:} Writing – review \& editing, Supervision, Methodology, Conceptualisation, Formal analysis. \textbf{Luis Jimenez-Linares:} Supervision,  Project administration, Methodology, Conceptualisation, Funding acquisition.

% --------------------
\section*{Declaration of competing interest}
The authors declare that they have no known competing financial interests or personal relationships that could have appeared to influence the work reported in this paper.

% --------------------
\section*{Data availability}
Data will be made available on request.

% --------------------
\section*{Acknowledgments}
This work was supported by grant PID2020-112967GB-C32 funded by MCIN/AEI/10.13039/501100011033 and by ERDF A Way of Making Europe. It was completed when Enrique Adrian Villarrubia-Martin was a predoctoral fellow at Universidad de Castilla-La Mancha funded by the European Social Fund Plus (ESF+) and in a visiting research stay at the University of Warwick funded by a mobility grant from the Universidad de Castilla-La Mancha for predoctoral students. Giovanni Montana acknowledges support from a UKRI AI Turing Acceleration Fellowship (EPSRC EP/V024868/1).

% --------------------
\appendix

\section{Transition Dynamics of the Simulation Kernel} \label{sec:transition_dynamics}

This section details the transition dynamics of the simulation kernel, which models the interactions between passenger decisions and operator strategies. The algorithm operates on a daily basis, initialising supply and demand modules to define available services and generate passenger demand. Valid journeys are filtered based on scheduling constraints and minimum transfer times, and utility values are calculated for each seat in the available journeys. Passengers choose the journey that maximises their utility, provided it is positive, or opt not to travel. The environment state is dynamically updated as tickets are purchased, reflecting the evolving interplay of passenger choices and pricing strategies. The detailed procedure of the simulation's transition dynamics of the environment is formalised in Algorithm~\ref{alg:transition_dynamics}.

\begin{algorithm}[!t]
    \caption{Transition Dynamics of the Simulation Kernel}
    \label{alg:transition_dynamics}
    \small
    \begin{algorithmic}[1]
        \Require Supply module $\mathcal{SM}$ with specified hyperparameters, demand module $\mathcal{DM}$ with specified hyperparameters, travel date $t$, minimum transfer time $\delta_{\text{min}}$
        \Ensure Updated environment with passenger decisions

        \State Initialise railway services $\mathcal{S}$ defined by operators in $\mathcal{SM}$
        \State Initialise user patterns $\mathcal{K}$ defined in $\mathcal{DM}$
        \State Generate passengers according to the daily demand $\mathcal{D}_t$ using Eq.~(\ref{eq:demand})
        
        \For{each passenger $n$ of type $k \in \mathcal{K}$}
            \State Filter valid journeys $\mathcal{J}$ for passenger market $w$ and desired travel date $t'$, ensuring minimum transfer time $\delta_{\text{min}}$
            \For{each journey $j \in \mathcal{J}$}
                \State Initialise journey utility $U_j \gets -\infty$
                \For{each service $s \in S_j$}
                    \State Initialise service utility $U_s \gets -\infty$
                    \For{each seat $c \in C_s$}
                        \If{no available tickets for seat $c$ on service $s$}
                            \State Set utility $U_{cs} \gets -\infty$
                        \Else
                            \State Set utility $U_{cs} \gets \delta_{ck} + \text{rand}(\varepsilon_k)$
                        \EndIf
                        \If{$U_{cs} > U_s$}
                            \State Update best seat $c^* \gets c$ and best utility $U_s \gets U_{cs}$
                        \EndIf
                    \EndFor
                \EndFor
                \If{$U_s > -\infty$}
                    \State Compute $U_j \gets U_{jc^*nt'}$ with the best seats $c^*$ using Eq.~(\ref{eq:utility})
                \EndIf
            \EndFor
            
            \If{$\max_{j \in \mathcal{J}} U_j > 0$}
                \State Select journey $j^*$ using Eq.~(\ref{eq:best_journey})
                \For{each service $s \in S_{j^*}$}
                    \State Buy a ticket for seat $c^*$ for service $s$ on journey $j^*$
                \EndFor
            \Else
                \State Passenger does not travel
            \EndIf
        \EndFor
    \end{algorithmic}
\end{algorithm}

\section{Hyperparameters} \label{sec:hyperparameters}

This section provides a comprehensive overview of the hyperparameters used to train the algorithms. For a standardised comparison, the default hyperparameters of each algorithm were used without additional tuning. Shared hyperparameters are presented in Table~\ref{tab:shared_hyperparameters}. Algorithm-specific configurations are detailed as follows: TD3 hyperparameters in Table~\ref{tab:td3_hyperparameters}, SAC, IQL-SAC, and VDN-SAC in Table~\ref{tab:sac_iql_vdn_sac_hyperparameters}, MAAC in Table~\ref{tab:maac_hyperparameters}, and MADDPG in Table~\ref{tab:maddpg_hyperparameters}.

\begin{table}[!t]
    \caption{Shared hyperparameters between algorithms.}
    \small
    \centering
    \begin{tabular}{lc}
        \toprule
        Hyperparameter & Value \\
        \midrule
        Number of environments & 16 \\
        Optimiser & Adam \cite{Kingma2015Adam:Optimization} \\
        Buffer size & 1,000,000 \\
        Training episodes & 200,000 \\
        Random policy episodes & 1,000 \\
        Evaluation episodes & 10,000 \\
        Number of hidden layers & 2 \\
        Hidden units per layer & 256 \\
        Non-linear activation function & ReLU \\
        Batch size & 256 \\
        Discount factor ($\gamma$) & 0.99 \\
        Target smoothing coefficient ($\tau$) & 0.005 \\
        Seed & 0, 43, 71 \\
        Training environment seed & $\{ \text{seed} + r \cdot 1000 \mid r \in \{0, 1, \ldots, 15 \}$ \\
        Evaluation environment seed & $\{ \text{seed} + r \cdot 100000 \mid r \in \{0, 1, \ldots, 15 \}$ \\
        Reward normalisation & Yes \\
        \bottomrule
    \end{tabular}
    \label{tab:shared_hyperparameters}
\end{table}

\begin{table}[!t]
    \caption{TD3 hyperparameters.}
    \small
    \centering
    \begin{tabular}{lc}
        \toprule
        Hyperparameter & Value \\
        \midrule
        Policy learning rate & 0.0001 \\
        Critic learning rate & 0.001 \\
        Hidden units per layer & [400, 300] \\
        Exploration noise & $\mathcal{N}(0, 0.1)$ \\
        Policy noise & $\mathcal{N}(0, 0.2)$ \\
        Policy noise clip & 0.5 \\
        Frequency delayed policy updates & 2 \\
        \bottomrule
    \end{tabular}
    \label{tab:td3_hyperparameters}
\end{table}

\begin{table}[!t]
    \caption{SAC, IQL-SAC, and VDN-SAC hyperparameters.}
    \small
    \centering
    \begin{tabular}{lc}
        \toprule
        Hyperparameter & Value \\
        \midrule
        Policy learning rate & 0.0003 \\
        Critic learning rate & 0.0003 \\
        Entropy regularisation coefficient ($\alpha$) & 0.2 \\
        Frequency delayed policy updates & 2 \\
        \bottomrule
    \end{tabular}
    \label{tab:sac_iql_vdn_sac_hyperparameters}
\end{table}

\begin{table}[!t]
    \caption{MAAC hyperparameters.}
    \small
    \centering
    \begin{tabular}{lc}
        \toprule
        Hyperparameter & Value \\
        \midrule
        Policy learning rate & 0.001 \\
        Critic learning rate & 0.001 \\
        Non-linear activation function & LeakyReLU \\
        Batch size & 1024 \\
        Number of attention heads & 4 \\
        Reward scale & 100 \\
        Steps per update & 100 \\
        Number of updates per update cycle & 4 \\
        \bottomrule
    \end{tabular}
    \label{tab:maac_hyperparameters}
\end{table}

\begin{table}[!t]
    \caption{MADDPG hyperparameters.}
    \small
    \centering
    \begin{tabular}{lc}
        \toprule
        Hyperparameter & Value \\
        \midrule
        Policy learning rate & 0.01 \\
        Critic learning rate & 0.01 \\
        Batch size & 1024 \\
        Discount factor ($\gamma$) & 0.95 \\
        Target smoothing coefficient ($\tau$) & 0.01 \\
        \multirow{2}{*}{Exploration noise} & Ornstein-Uhlenbeck \cite{Uhlenbeck1930OnMotion} \\
        & $\theta = 0.15$ $\sigma = 0.2$ \\
        \bottomrule
    \end{tabular}
    \label{tab:maddpg_hyperparameters}
\end{table}

\clearpage

\bibliographystyle{elsarticle-num}
\bibliography{references}
\end{document}